\documentclass[journal]{IEEEtran}

\usepackage[numbers,sort&compress]{natbib} 
\usepackage{float}

\usepackage{multirow}
\usepackage{threeparttable}

\usepackage[ruled,linesnumbered]{algorithm2e} 
\usepackage{amsmath}
\usepackage{amsfonts}
\usepackage{amssymb}  

\usepackage{graphicx}
\graphicspath{{images/}}                     

\usepackage{hyperref}

\usepackage[utf8]{inputenc}

\usepackage{pifont}

\usepackage{booktabs}

\usepackage{soul}
\usepackage{color, xcolor}
\soulregister\cite7
\soulregister\citep7
\soulregister\citet7 
\soulregister\ref7 
\soulregister\pageref7 
\usepackage{xcolor}
\usepackage{colortbl}

\definecolor{myyellow}{RGB}{255,255,153} 
\definecolor{myred}{RGB}{255,153,153}   
\definecolor{myorg}{RGB}{255,165,0} 
\definecolor{mywhite}{RGB}{255,255,255}

\usepackage{makecell}

\usepackage{dcolumn}
\newcolumntype{C}{>{\arraybackslash}c} 
\newcolumntype{D}{D{/}{/}{5}}

\begin{document}
%


\title{FootprintNet: State-Transition-Guided Dynamic Footprint Learning for Multi-temporal Remote Sensing Change Detection}


\author{Haotian Zhang$^{1}$, Hao Chen$^{2}$, Han Guo$^{1}$, Zhengxia Zou$^{1}$, and Zhenwei Shi$^{1, \star}$ 
\\
\vspace{6pt}
Beihang University$^1$, Shanghai Artificial Intelligence Laboratory$^2$

}
\date{April. 2023}

\maketitle
\begin{abstract}


Despite substantial progress in remote sensing multi-temporal change detection (MTCD), most existing MTCD methods still represent the dynamic process at each spatial location over the entire observation period using a single change category associated with the final observation. This implicit single-change assumption limits their ability to characterize regions of recurrent change closely related to human activities. To address this limitation, we introduce Urban Building Dynamics Detection (UBDD), which identifies building-change dynamic footprints, i.e., the temporal intervals in which changes occur, from multi-temporal imagery and produces pixel-wise classification masks. For regions undergoing two or more changes, UBDD introduces an independent multi-change class for unified representation, thereby enabling unified modeling of single- and multi-change processes. Furthermore, we propose FootprintNet, which abstracts building-change processes as interactions between latent states and actions, and imposes state-action transition constraints to guide the learning of causally coherent change trajectories. It further exploits temporal change-boundary cues to enhance feature contrast across boundary sides, thereby improving the discrimination among different dynamic footprints and enabling accurate detection of dynamic footprints. Moreover, we introduce the Building Change Dynamics Score (BCDS) to address the inability of conventional metrics to reflect the temporal proximity between predicted footprints and labels. It evaluates predictions according to their preservation of change semantics and temporal offsets from the corresponding labels. Extensive experiments on TSCD, MUDS, and WUSU demonstrate that FootprintNet outperforms current state-of-the-art methods, achieving mIoU improvements of 0.67\%, 1.12\%, and 5.84\%, respectively. The code is available at https://github.com/zmoka-zht/FootprintNet.
\end{abstract}


\IEEEpeerreviewmaketitle


\section{Introduction}
\label{sec:intro}

\IEEEPARstart{C}{hange} detection is a fundamental Earth observation task that identifies land-cover changes from remote sensing images acquired at different times. It has significant applications in urban construction monitoring~\cite{coppinp2004digitalchangedetection,11370463, 11550391, 10019593}, disaster damage assessment~\cite{zheng2021building, 11345538, 10950382, 11217335}, and land resource management~\cite{li2016superresolution,lu2004change,9891887,10769517}. With the continuous advancement of Earth observation technologies, the availability of multi-temporal remote sensing data has increased substantially, enabling continuous and fine-grained monitoring of the same region. Beyond determining whether changes occur, identifying when they occur provides deeper insights into regional development and the dynamic patterns of surface objects.

However, existing remote sensing change detection studies predominantly focus on bi-temporal change detection, which identifies changed and unchanged regions by comparing remote sensing images acquired at two different times~\cite{10599227, SAAN,lei2021difference,feng2022icif,BIT, zhang2024bifa, zhang2025cdmamba, zhang2025foba, 10591792, guo2025taco, 10287990, 11614676, 9975266, 11489274}. Nevertheless, bi-temporal change detection retains only the surface states at the beginning and end of the observation period, essentially capturing the accumulated changes between two observations. When the temporal interval is large, intermediate change information and its underlying dynamics may be substantially compressed or even completely lost, making it difficult to characterize continuous or recurrent changes occurring throughout the observation period.

To alleviate the temporal information loss inherent in bi-temporal change detection, increasing attention has been devoted to multi-temporal change detection (MTCD). Given multi-temporal observations of the same region, this task identifies the spatial locations and occurrence times of changes by either independently modeling multiple adjacent temporal pairs~\cite{hafner2025continuous,zhao2024coud,qu2026anytimecd,kennedy2010detecting, he2024time, zhu2014continuous} or jointly modeling the entire multi-temporal sequence~\cite{li2024difference,CAIMNet}. Compared with bi-temporal change detection, multi-temporal change detection localizes surface changes at a higher temporal resolution and provides finer-grained temporal information for understanding regional development dynamics.

Nevertheless, most existing methods remain on a single change time and typically represent the dynamic process at each spatial location over the entire observation period using only the category associated with its last change time. This formulation implicitly assumes that each location undergoes at most one change, making it difficult to characterize regions experiencing continuous or recurrent processes, such as construction-demolition-reconstruction. Such regions are closely associated with intensive human activities, including urban renewal and post-disaster reconstruction, and are therefore of considerable importance for dynamic urban construction monitoring.

To address the above limitations, we extend existing multi-temporal change detection and introduce Urban Building Dynamics Detection (UBDD), which aims to characterize the dynamic footprints of building changes from multi-temporal observations. Here, dynamic footprints refer to the temporal intervals during which different building regions change over the complete observation period. Given a sequence of remote sensing images acquired over the same region at different times, UBDD jointly determines whether a building has changed and identifies when the change occurred, producing a pixel-wise classification mask in which different categories correspond to different dynamic footprints. For locations undergoing two or more changes during the complete observation period, UBDD does not recover the exact occurrence time of each individual event. Instead, it introduces an independent multi-change category to provide a unified representation. Consequently, UBDD retains the capability to identify single-change footprints while relaxing the implicit single-change assumption of existing MTCD, thereby enabling unified modeling of unchanged, single-change, and multi-change processes and providing a more complete representation of building dynamics.

As UBDD targets the continuous observation of real-world objects over time, the construction, demolition, and reconstruction of artificial structures, particularly buildings, generally follow specific physical state-transition rules. Meanwhile, compared with binary or semantic change detection based on bi-temporal imagery, UBDD involves longer observation sequences and is therefore more susceptible to irrelevant factors, such as illumination variations and seasonal differences, making it difficult to learn stable and discriminative temporal representations.

To address these challenges, we propose FootprintNet, which consists of three complementary branches. First, the Action-Guided Latent State Transition branch (ALST) abstracts building-change processes as interactions between latent states and actions, and imposes state-action transition constraints to guide the model in learning causally coherent change trajectories. Second, the Asymmetric Temporal-Boundary Contrastive branch (ATBC) treats change times as temporal boundaries and enhances the feature discrepancy across the two sides of each boundary, thereby improving the discrimination of different dynamic footprints while suppressing interference from irrelevant factors. Finally, the Spatial-Temporal State Space Scanning branch (STSS) models spatial-temporal dependencies across consecutive observations, enabling accurate prediction of the dynamic footprints of building changes.

Moreover, conventional pixel-wise classification metrics typically impose equivalent penalties on prediction errors across different categories, making it difficult to reflect the varying impacts of different error types on the representation of building-change dynamic footprints in UBDD. Since UBDD concerns not only the spatial localization of changed regions but also the preservation of temporal semantics throughout the entire observation period, we further introduce the Building Change Dynamics Score (BCDS). This metric assigns differentiated scores according to the temporal semantic consistency between the predicted and label, thereby providing higher evaluations for predictions that better preserve temporal semantics. Meanwhile, it explicitly measures dynamic footprint offsets among different single-change categories to characterize change-time localization errors, enabling a more comprehensive evaluation of the predicted dynamic footprints of building changes.

In summary, the main contributions of this paper can be summarized as follows:
\begin{itemize}
\item \textbf{Task level:} We extend existing MTCD and introduce Urban Building Dynamics Detection (UBDD). UBDD generalizes conventional single-change-time localization to the unified identification of single and recurrent changes, relaxing the implicit assumption of a single change process.

\item \textbf{Method level:} We propose FootprintNet, which explicitly models state-action transition rules of building changes and enhances feature discrimination among different dynamic footprints, thereby enabling accurate prediction of building-change dynamic footprints.

\item \textbf{Evaluation level:} We introduce the Building Change Dynamics Score (BCDS), which assigns differentiated scores according to temporal semantic preservation and change-time offsets. By overcoming the limitation of conventional metrics that penalize different semantic errors equally, BCDS provides a more comprehensive assessment of the prediction quality of building dynamics.

\item \textbf{Experimental level:} Extensive quantitative and qualitative experiments on three datasets demonstrate the effectiveness of the proposed method.
\end{itemize}

The remainder of this paper is organized as follows: Section \ref{sec:relatedwork} reviews related work. Section \ref{sec:method} provides a detailed description of the proposed FootprintNet. Some experimental results are reported in Section \ref{sec:experiment}. The conclusion is made in Section \ref{sec:conclusion}.


\section{Related Work} \label{relatedwork}
\label{sec:relatedwork}

\subsection{Multi-temporal Change Detcetion}
Existing multi-temporal change detection methods can be broadly divided into three categories. The first extends bi-temporal change detection to multi-temporal inputs while predicting the accumulated changes between the first and last observations. Representative recurrent approaches include L-UNet~\cite{9352207}, which embeds fully convolutional LSTM modules into a U-Net-like encoder, and MC2ABNet~\cite{10285305}, which employs ConvBiLSTM to capture bidirectional temporal dependencies. For irregularly sampled sequences, Yang et al.~\cite{9771449} introduced a time-distance-guided LSTM that adaptively adjusts its input and forget gates according to temporal intervals. Beyond RNN architectures, Meshkini et al.~\cite{10416178} adopted pretrained 3D CNNs to extract joint spatiotemporal features and later extended this paradigm to weakly supervised change detection. More recently, Zhao et al.~\cite{zhao2025histenet} introduced state space models to efficiently capture complex spatiotemporal correlations in multi-temporal imagery.

The second category improves temporal resolution by detecting changes between adjacent observations rather than only the accumulated changes between the first and last images. Representative methods emphasize continuous multi-temporal representation learning. Hafner et al.~\cite{hafner2025continuous} proposed a multi-task framework with a self-attention based temporal refinement module to preserve temporal information and enhance sensitivity to continuous changes. Zhao et al.~\cite{zhao2024coud} introduced COUD, which combines self-supervised pretraining with temporal distillation to strengthen feature representations. Similarly, AnyTime-CD~\cite{qu2026anytimecd} jointly models temporal, spectral, and spatial information through multi-task learning, thereby improving the characterization of complex temporal change processes.

The third category can be regarded as a semantic extension of the preceding two paradigms. These methods typically adopt a dual-branch output architecture, where one branch predicts the change mask between the first and last observations, while the other identifies the change time. Li et al.~\cite{li2024difference} pioneered this direction by integrating optical flow with the LSTM to model change differences across multi-temporal images. Building upon this work, they further proposed CAIM-Net~\cite{CAIMNet}, which employs a coarse-to-fine change-time extraction strategy to improve the consistency between change-region localization and change-time prediction.

Distinct from the aforementioned TCD paradigms, the proposed UBDD retains the capability to identify single-change temporal footprints while relaxing the implicit single-change assumption of existing TCD methods, thereby providing a unified representation of unchanged, single-, and multi-change processes.

\begin{figure*}
    \centering
    \includegraphics[width=0.99\textwidth]{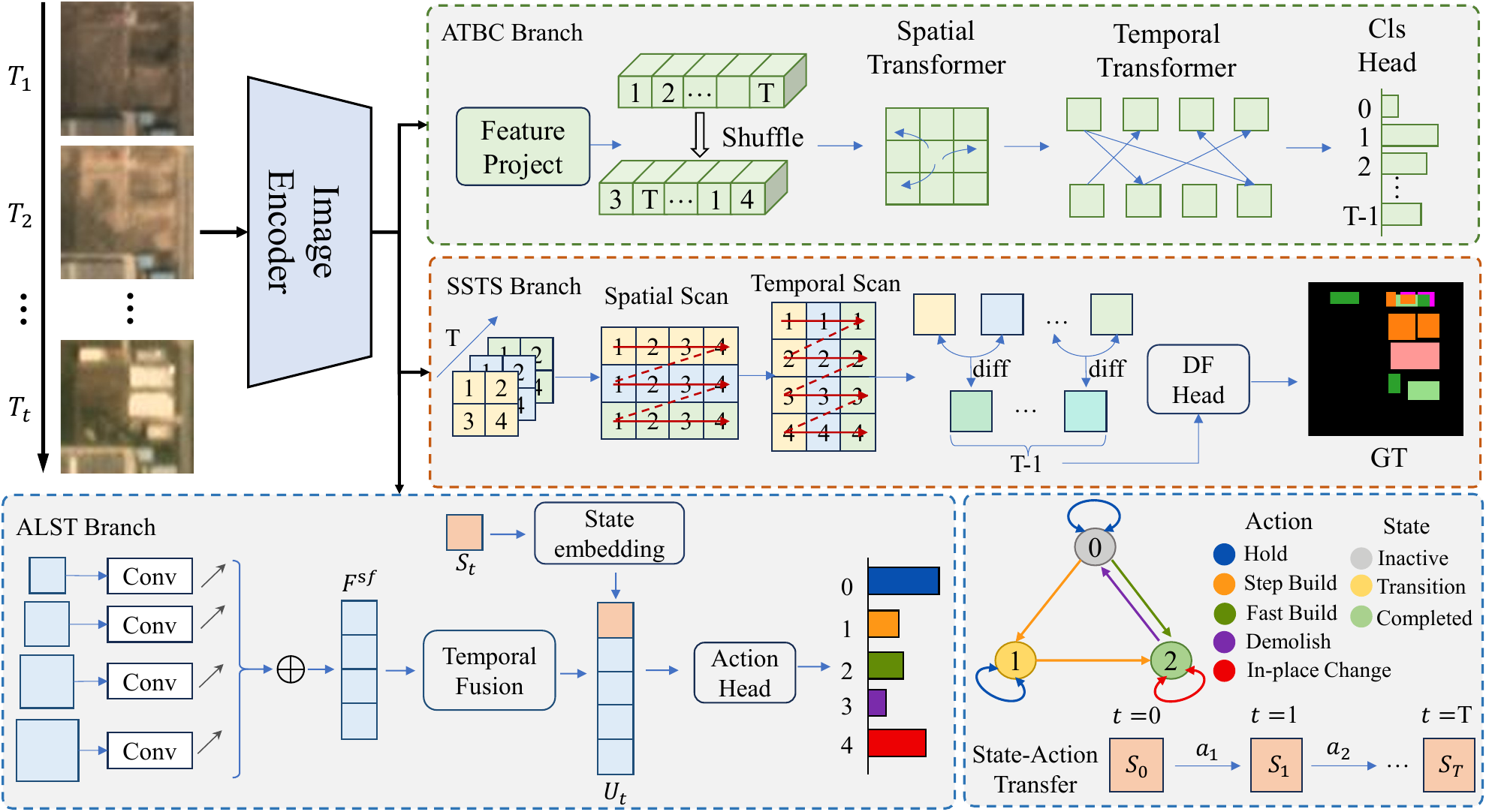}
    \caption{
    Overall architecture of the proposed method. ALST, ATBC, and STSS denote the Action-guided Latent State Transition branch, the Asymmetric Temporal-Boundary Contrastive branch, and the Spatial-Temporal State Space branch, respectively. The $\nearrow$ indicate upsampling operations, while DF Head denotes the dynamic footprint prediction head.
    }
    \label{fig:FootprintsNet}
\end{figure*}

\subsection{Reinforcement Learning in Remote Sensing Change Detection}
Recently, several studies have introduced reinforcement learning (RL) into remote sensing image change detection~\cite{9883633, 10605273, 9956544, sahbi2023reinforcement, pujari2026deep, ma2025vilacd}. Deschamps et al.~\cite{9883633} developed an RL-based active learning strategy to improve model generalization. Pujar et al.~\cite{pujari2026deep} proposed the Deep Gated Hyperbolic Sine Reinforcement with Q-learning (DGHSQ) for semantic change detection, while Ma et al.~\cite{ma2025vilacd} combined supervised fine-tuning and reinforcement learning to train a vision-language model for coarse change-mask generation. Zhao et al.~\cite{zhao2026optimized} formulated change-probability refinement as a Markov process and employed PPO~\cite{schulman2017proximal} for iterative pixel-wise optimization. In addition, RL has also been applied to upstream image augmentation. Huang et al.~\cite{10714383} proposed ADAAUG+, an adaptive data augmentation method that dynamically selects effective augmentation policies for individual training samples.

Unlike existing RL-based change detection methods, FootprintNet employs RL to model building dynamics by characterizing state-action transition trajectories across temporal observations.


\section{FootprintNet}
\label{sec:method}

\subsection{Overview}
\label{ssec:overview}
Urban Building Dynamics Detection aims to identify the dynamic footprints of building changes from multi-temporal images. To this end, we propose FootprintNet, which comprises three complementary branches: the Action-guided Latent State Transition branch (ALST), the Asymmetric Temporal-Boundary Contrastive branch (ATBC), and the Spatial-Temporal State Space branch (STSS), as illustrated in Fig.~\ref{fig:FootprintsNet}. Given an image sequence $\{{I_t}\}_{t=1}^{T}$, where $I_t \in \mathbb{R}^{C \times H \times W}$, $C$, $H$, $W$, $T$ denote the number of channels, image height, image width, and sequence length, respectively. Each image is processed by a weight-shared encoder to extract four-level features $\{\mathbf{F}_t^i\}_{i=1}^{4}$.


For the ALST branch, features from four encoder stages are aggregated to provide the contextual information required for state-action modeling. Multi-scale fusion first integrates different semantic levels, yielding $\{{F^{sf}_t}\}_{t=1}^{T} \in \mathbb{R}^{B \times C_{sf} \times H \times W}$. Temporal interactions are then performed to obtain $F_{tf} \in \mathbb{R}^{B \times T \times C_{tf} \times H \times W}$. To model the state-transition dynamics, we define three discrete states for each spatial location at different time steps: Inactive ($0$), Transition ($1$), and Completed ($2$). The current state is encoded as $\mathbf{F}^{\mathrm{state}}_{t}\in\mathbb{R}^{B\times C_s\times H\times W}$. We formulate the sequential state-action evolution as a Markov process, where the action at time step $t$ is conditioned on $\mathbf{F}_{tf}^{t}$ and $\mathbf{F}_{\mathrm{state}}^{t}$, while the next state is determined by the resulting state-action pair. Subsequently, their fused representation is fed into an action prediction head to estimate the transition between adjacent observations, producing $\mathbf{P}^{action}_t \in \mathbb{R}^{B \times C_a \times H \times W}$. Finally, GRPO~\cite{shao2024deepseekmath} is employed to constrain the state-action evolution toward consistent real-world building-change process.

In the ATBC branch, only the last-stage encoder features are projected and reshaped to obtain $\mathbf{F}_p\in\mathbb{R}^{B\times T\times N\times C}$, providing high-level semantic representations. These features are randomly shuffled along the temporal dimension and rearranged as ${\mathbf{F}}_{shuf}\in\mathbb{R}^{BT\times N\times C}$. A spatial Transformer first captures intra-temporal dependencies among patches, after which the features are reshaped into $\mathbb{R}^{BN\times T\times C}$ and processed by a temporal Transformer to model cross-temporal interactions. Finally, a classification head produces $\mathbf{P}_{con}\in\mathbb{R}^{B\times T\times N\times C_T}$ to predict the temporal stage of each patch. Based on these predictions, the proposed Asymmetric Temporal-Boundary Loss constrains patch-level representations to enhance their sensitivity to temporal change boundaries.

In the STSS branch, we construct an SSM~\cite{gu2024mamba} based temporal footprint prediction head. By exploiting scanning, STSS jointly models spatial dependencies within each inter-temporal and cross-temporal interaction, thereby capturing the spatiotemporal correlations of building changes. Progressive upsampling and multi-scale fusion are then applied to aggregate features across encoder stages, yielding$\{\mathbf{F}^{ssst}_t\}_{i=1}^T \in \mathbb{R}^{B \times C \times H \times W}$. Adjacent temporal features are further differenced to obtain $\mathbf{F}_{diff} \in \mathbb{R}^{B \times (T-1) \times C_{diff}\times H \times W}$, which is finally fed into the prediction head to generate building-change dynamics footprints.


\subsection{Action-Guided Latent State Transition Branch}
\label{ssec:ALST}
The ALST branch formulates sequential state-action evolution as a Markov process, explicitly incorporating building evolution rules to guide state-transition learning, as illustrated in Fig.~\ref{fig:FootprintsNet}. At each time step, pixel-level temporal features are combined with the current state to predict an action, which updates the next state according to predefined transition rules, thereby constructing the complete state evolution trajectory.

Given the multi-level temporal features $\{\mathbf{F}_t^i\}_{i=1}^{4}$, a multi-scale fusion module first aligns all features to $1/4$ of the input resolution. The aligned features are then concatenated along the channel dimension and projected to integrate information across different levels. To support pixel-wise state and action prediction, the fused features are upsampled to the original resolution, yielding  $\{{F^{sf}_t}\}_{t=1}^{T}$.
To enable cross-temporal interaction, the features from different temporal are aggregated and fused along the channel dimension using convolution. The resulting feature is then reshaped into $F^{tf} \in \mathbb{R}^{B \times T \times C_{tf} \times H \times W}$ for subsequent state modeling. 

For each pixel, we define a state $\mathbf{S}_{t}\in{0,1,2}^{B\times H\times W}$ to represent its status at time step $t$. The initial state is inferred from the image features at $t=0$. During training, the temporal feature $\mathbf{F}^{tf}_t$ is replicated across $G$ parallel exploration trajectories, yielding $\mathbf{F}^{tf}_t \in\mathbb{R}^{GB\times C_{r}\times H\times W}$~\cite{shao2024deepseekmath}. Meanwhile, the current state $\mathbf{S}_{t}$ is encoded by a state embedding layer to obtain $\mathbf{E}_{t}$. The expanded temporal feature and state embedding are then fused into the joint representation $\mathbf{U}_{t}$:
\begin{equation}
\label{eq:state_embedding_fusion}
\begin{aligned}
\mathbf{E}_{t}
&=
\Phi_{e}(\mathbf{S}_{t}),
\\
\mathbf{E}_{t}^{g}
&=
\operatorname{Repeat}_{G}(\mathbf{E}_{t}),
\\
\mathbf{U}_{t}
&=
\Phi_{f}
\left(
\mathbf{F}^{tf}_t,
\mathbf{E}_{t}^{g}
\right)
\end{aligned}
\end{equation}
where, $\Phi_{e}(\cdot)$ denotes the state embedding layer, $\operatorname{Repeat}_{G}(\cdot)$ denotes the operation of replicating the feature $G$ times along the batch dimension to match the parallel exploration trajectories, and $\Phi_{f}(\cdot)$ denotes the feature fusion mapping function. 

The fused feature $\mathbf{U}_{t}$ is then fed into the action head to predict the action executed for transitioning from the current state to the next:
\begin{equation}
\label{eq:pixel_action_prediction}
\begin{aligned}
\boldsymbol{\pi}_{t}(p) 
&= \operatorname{Softmax}\left(\Phi_{a}\big(\mathbf{U}_{t}(p)\big)\right), \\
a_{t}(p) 
&\sim \operatorname{Categorical}\big(\boldsymbol{\pi}_{t}(p)\big)
\end{aligned}
\end{equation}
where, $a_t(p) \in \mathcal{A}$ and $\mathcal{A}=\{0,1,2,3,4\}$, $\Phi_{a}(\cdot)$ denotes the action prediction head, $\boldsymbol{\pi}_{t}$ denotes the action probability distribution, and $a_{t}$ denotes the action sampled from this distribution. $a_{0}$, $a_{1}$, $a_{2}$, $a_{3}$, and $a_{4}$ denote hold, step build, fast build, demolish, and in-place change, respectively. 

After obtaining the sampled action, the next state $\mathbf{S}_{t+1}$ is updated from the current state $\mathbf{S}_{t}$ and action $a_{t}$. The state transition process can be defined as:
\begin{equation}
    \mathbf{S}_{t+1}=\mathcal{T}(\mathbf{S}_{t},a_{t})
\end{equation}
where, $\mathcal{T}(\cdot)$ denotes the predefined state transition function. 

Specifically, $a_{0}$ leaves the current state unchanged; $a_{1}$ gradually increases the state, while its maximum value is constrained by the completed change state, i.e., $\mathbf{S}_{t}=2$; $a_{2}$ is only allowed when $\mathbf{S}_{t}=0$ and directly drives the state to the completed change state; $a_{3}$ resets the current state to $0$; and $a_{4}$ keeps the state at $2$. These transition rules are summarized in Algorithm~\ref{alg:state_transition}.

\begin{algorithm}[t]
\caption{State Transition Rule}
\label{alg:state_transition}
\KwIn{Current state $s_t \in \{0,1,2\}$ and action $a_t \in \{0,1,2,3,4\}$}
\KwOut{Next state $s_{t+1}$}

$s_{t+1} \leftarrow s_t$\tcp*[r]{Default transition}

\uIf{$a_t = 0$}{
    $s_{t+1} \leftarrow s_t$\tcp*[r]{Hold}
}
\uElseIf{$a_t = 1$}{
    $s_{t+1} \leftarrow \min(s_t+1,2)$\tcp*[r]{Step Build}
}
\uElseIf{$a_t = 2$ and $s_t = 0$}{
    $s_{t+1} \leftarrow 2$\tcp*[r]{Fast Build}
}
\uElseIf{$a_t = 3$}{
    $s_{t+1} \leftarrow 0$\tcp*[r]{Demolition}
}
\uElseIf{$a_t = 4$}{
    $s_{t+1} \leftarrow 2$\tcp*[r]{In-place change}
}

\Return{$s_{t+1}$}
\end{algorithm}

The above process only describes the state transition at time step $t$. Recursively unrolling this transition over the temporal dimension yields the complete state and action trajectories across the $T$ observations.

\textbf{Action-Guided Latent State Transition Reward:} To learn state-action trajectories consistent with real temporal dynamics, we introduce a transition reward conditioned on the predicted state trajectory, action sequence, and labels. The hybrid trajectory-level reward comprises validity, label-consistency, temporal-consistency, and multi-change terms.


Specifically, given $T$ temporal images, we denote the number of dynamic footprints by $L=T-1$. For pixel $i$, $y_i=0$ denotes the unchanged class, $y_i\in\mathcal{C}_s=\{1,\ldots,L\}$ denotes a single-change class, and $y_i=c_m$ denotes the multi-change class. The trajectory-level statistics used below are described as follows. 

Across the $L$ dynamic footprints, $N_i^{\mathrm{inv}}$ denotes the number of invalid actions that violate the state-action transition rules defined in Algorithm~\ref{alg:state_transition}, whereas $N_i^{\mathrm{fake}}$ counts the \textbf{in-place} actions performed in unchanged regions.  $N_i^{\mathrm{act}}$ denotes the total number of non-\textbf{hold} actions, while $N_i^{\mathrm{vact}}$ counts the valid non-\textbf{hold} actions that satisfy the prescribed transition rules. For a single-change pixel, $N_i^{\mathrm{out}}$ represents the number of non-\textbf{hold} actions executed outside the ground-truth dynamic footprints. Moreover, $\hat{z}_i$ denotes the trajectory-level change indicator, which is set to 1 when $N_i^{\mathrm{vact}}>0$ and to 0 otherwise. The indicator $h_i^{\mathrm{tar}}$ determines whether the ground-truth dynamic footprint is correctly activated: $h_i^{\mathrm{tar}}=1$ if at least one valid non-\textbf{hold} action occurs within that interval, and 0 otherwise. The clean-trajectory indicator $\mathcal{C}_i$ evaluates whether the entire trajectory contains neither invalid actions nor spurious \textbf{in-place} actions. It is set to 1 when both conditions are satisfied and to 0 otherwise. The formula is as follows.

The validity reward $R_{\mathrm{valid}}$ enforces the transition rules in Algorithm~\ref{alg:state_transition} and penalizes invalid state-action transitions:
\begin{equation}
\begin{aligned}
R_i^{\mathrm{val}}
={}
&+0.4\,\mathbb{I}\!\left[
y_i\neq0,\,
\hat{z}_i=1,\,
\mathcal{C}_i=1
\right] \\ 
&+0.3\,\mathbb{I}\!\left[
y_i=0,\, 
N_i^{\mathrm{act}}=0,\,
\hat{z}_i=0,\, 
\mathcal{C}_i=1 
\right] \\
&-2N_i^{\mathrm{fake}} \\
&-3N_i^{\mathrm{inv}}
\end{aligned} \label{eq:validity_reward} \end{equation}

The label-consistency reward $R_{\mathrm{label}}$ evaluates whether the predicted trajectory agrees with the changed and unchanged regions defined by the labels:
\begin{equation}
\begin{aligned} 
R_i^{\mathrm{lab}} 
={}
&+2\,\mathbb{I}\!\left[
y_i=0,\, 
N_i^{\mathrm{act}}=0,\, 
\hat{z}_i=0 
\right] \\ 
&+\mathbb{I}\!\left[ 
y_i\neq0,\, 
\hat{z}_i=1 
\right] \\ 
&-\mathbb{I}\!\left[ 
y_i=0,\, 
\hat{z}_i=1 
\right] \\ 
&-3\,\mathbb{I}\!\left[
y_i=0,\, 
N_i^{\mathrm{act}}>0 
\right] \\ 
&-3\,\mathbb{I}\!\left[ 
y_i\neq0,\, 
\hat{z}_i=0 
\right] 
\end{aligned} \label{eq:label_consistency_reward} \end{equation}

For single-change regions, the temporal-consistency reward $R_{\mathrm{temporal}}$ encourages valid actions within the annotated change interval while suppressing redundant actions elsewhere:
\begin{equation}
\begin{aligned}
R_i^{\mathrm{temporal}}
={}&
+1.8\,\mathbb{I}\!\left[
y_i\in\mathcal{C}_s,\,
h_i^{\mathrm{tar}}=1,\,
N_i^{\mathrm{out}}=0
\right]
\\
&-1.2\,\mathbb{I}\!\left[
y_i\in\mathcal{C}_s
\right]N_i^{\mathrm{out}} \\
&-2.5\,\mathbb{I}\!\left[
y_i\in\mathcal{C}_s,\,
h_i^{\mathrm{tar}}=0
\right]
\end{aligned}
\label{eq:temporal_consistency_reward}
\end{equation}

For multi-change regions, $R_{\mathrm{multi}}$ requires valid transitions across multiple change intervals and penalizes trajectories that collapse into a single evolution event:
\begin{equation} 
\begin{aligned} 
R_i^{\mathrm{multi}} ={}
&+1.8\,\mathbb{I}\!\left[
y_i=c_m,\, 
N_i^{\mathrm{vact}}\geq2 
\right] \\
&-1.5\,\mathbb{I}\!\left[
y_i=c_m,\,
N_i^{\mathrm{act}}\geq2,\, N_i^{\mathrm{vact}}<2 
\right] \\
&-1.8\,\mathbb{I}\!\left[
y_i=c_m,\, 
N_i^{\mathrm{vact}}=1 
\right] \\
&-2\,\mathbb{I}\!\left[
y_i=c_m,\, 
\hat{z}_i=0 
\right] \\
&-3\,\mathbb{I}\!\left[
y_i=c_m,\,
N_i^{\mathrm{vact}}=0
\right] 
\end{aligned} \label{eq:multi_change_reward} \end{equation}

Finally, the above rewards are weighted and accumulated to obtain the hybrid reward score:
\begin{equation}
\label{eq:overall_reward}
R_{\mathrm{all}}
=
R_{\mathrm{valid}}
+
R_{\mathrm{label}}
+
R_{\mathrm{temporal}}
+
R_{\mathrm{multi}}
\end{equation}

\subsection{Asymmetric Temporal-Boundary Contrastive Branch}
\label{ssec:ATBC}
The ATBC branch explicitly exploits temporal change-boundary cues to enhance feature discrimination between different temporal-dynamics regions, promoting clearer separation in the feature space.

As shown in Fig.~\ref{fig:FootprintsNet}, the ATBC branch takes the last-stage encoder feature $\mathbf{F}_{4}$ as input. It is first fed into the feature project module consisting of layer normalization, linear projection, and GELU activation to produce $\mathbf{F}_{p} \in \mathbb{R}^{B \times T \times N \times C_p}$ ($N$ is $H_4 \times W_4$). To reduce reliance on fixed temporal positions and enhance sensitivity to change boundaries, $\mathbf{F}_{p}$ is randomly shuffled along the temporal dimension, yielding $\mathbf{F}_{shuf}$. Subsequently, $\mathbf{F}_{shuf}$ is fed into the spatial Transformer, which captures global dependencies among patches within each temporal, followed by the temporal Transformer that models contextual interactions inter-temporally, producing $\mathbf{F}_{st}$. Finally, the classification head predicts the temporal stage of each patch. This process is formulated as:
\begin{equation}
\label{eq:atbc_forward}
\begin{aligned}
\mathbf{F}_{p} &= \Phi_{p}(\mathbf{F}_{4}), \\
\mathbf{F}_{shuf} &= \mathcal{S}_{T}(\mathbf{F}_{p}), \\
\mathbf{F}_{st} &= \Phi_{t}\big(\Phi_{s}(\mathbf{F}_{shuf})\big), \\
\mathbf{P}_{con} &= \Phi_{cls}(\mathbf{F}_{st})
\end{aligned}
\end{equation}
where, $\Phi_{p}(\cdot)$ denotes the feature project module composed of layer normalization, linear projection, and a GELU activation function. $\mathcal{S}_{T}(\cdot)$ denotes the random shuffling operation along the temporal dimension. $\Phi_{s}(\cdot)$ and $\Phi_{t}(\cdot)$ denote the spatial Transformer and temporal Transformer, respectively. And $\Phi_{cls}(\cdot)$ denotes the classification head.

\textbf{Asymmetric Temporal-Boundary Loss:} To enhance feature separation between different temporal-dynamics regions, we introduce the Asymmetric Temporal-Boundary Loss. During training, a state-mapping matrix $\mathbf{M}_{s}$ is produced from the labels $\mathbf{Y}$ to encode different temporal-dynamics categories.


Taking a four-frame sequence as an example, the unchanged class is encoded as $[0,0,0,0]$, whereas a change occurring between $T_1$ and $T_2$ is represented as $[0,1,1,1]$. The remaining single-change classes are encoded analogously. The multi-change class is represented as $[0,1,2,3]$. These state mappings serve as temporal supervision for the classification head and are optimized using cross-entropy loss.

Meanwhile, a boundary-mapping matrix $\mathbf{M}_{b}$ is constructed to explicitly encode change boundaries between adjacent temporal frames. Each change category is assigned according to its boundary location. For example, a change between $T_1$ and $T_2$ is represented as $[0,1,0]$.
\begin{equation}
\label{eq:state_boundary_mapping}
\begin{aligned}
\mathbf{M}_{s} &= \Psi_{s}(\mathbf{Y}),
& \mathbf{M}_{s} &\in \mathbb{R}^{B \times N \times T}, \\
\mathbf{M}_{b} &= \Psi_{b}(\mathbf{Y}),
& \mathbf{M}_{b} &\in \mathbb{R}^{B \times N \times (T-1)}
\end{aligned}
\end{equation}
where $\Psi_s(\cdot)$ and $\Psi_b(\cdot)$ denote the mapping function generates the state mapping matrix and the boundary mapping matrix from the labels.

In addition, to ensure consistency with the temporal feature shuffling in the ATBC branch, $\mathbf{M}_{s}$ and $\mathbf{M}_{b}$ are synchronously shuffled using the same permutation indices, thereby preserving their temporal alignment with the shuffled features.
\begin{equation}
\label{eq:shuffle_state_boundary_mapping}
\widetilde{\mathbf{M}}_{s}
=
\mathcal{S}_{T}\left(\mathbf{M}_{s}\right),
\quad
\widetilde{\mathbf{M}}_{b}
=
\mathcal{S}_{T}\left(\mathbf{M}_{b}\right)
\end{equation}

To emphasize responses at temporal-dynamics boundaries, we introduce an asymmetric weighting strategy. Specifically, temporal-dynamics boundary patches are assigned larger weights to strengthen discriminative learning, whereas unchange patches receive smaller weights to mitigate the dominance of abundant unchanged regions during optimization.


\subsection{Spatial-Temporal State Space Branch}
\label{ssec:STSS}
The SSTS branch aims to models spatiotemporal correlations across sequential images to accurately predict dynamic footprints. To this end, we design a SSM-based prediction head that captures both spatial dependencies and temporal interactions. 

Given the multi-scale temporal features ${\{\mathbf{F}_{t}^{i}\}}_{i=1}^{4}$ extracted by the encoder, the temporal index is omitted for simplicity. The smallest scale feature $\mathbf{F}^{4}$ is first rearranged by merging the batch and temporal dimensions, and then processed by the spatial SSM to capture intra-frame dependencies, yielding $\mathbf{F}_{spatial}^{4} \in \mathbb{R}^{BT \times C_{4} \times H_{4} \times W_{4}}$. Subsequently, the spatial dimensions are flattened and merged with the batch dimension, after which the temporal SSM models cross-temporal dependencies at each spatial location, producing $\mathbf{F}_{temporal}^{4} \in \mathbb{R}^{BH_{4}W_{4} \times C_{4} \times T \times 1}$. This process is formulated as:
\begin{equation}
\label{eq:ssts_scanning}
\begin{aligned}
\mathbf{F}_{\mathrm{spatial}}^{4}
&=
\Phi_{\mathrm{SSM}}^{s}
\left(
\mathcal{R}_{s}\left(\mathbf{F}^{4}\right)
\right), \\
\mathbf{F}_{\mathrm{temporal}}^{4}
&=
\Phi_{\mathrm{SSM}}^{t}
\left(
\mathcal{R}_{t}\left(\mathbf{F}_{\mathrm{spatial}}^{4}\right)
\right)
\end{aligned}
\end{equation}
where, $\mathcal{R}_{s}(\cdot)$ and $\mathcal{R}_{t}(\cdot)$ denote the feature rearrangement operations for spatial scanning and temporal scanning, respectively, while $\Phi_{\mathrm{SSM}}^{s}(\cdot)$ and $\Phi_{\mathrm{SSM}}^{t}(\cdot)$ denote the spatial SSM and the temporal SSM, respectively.

Subsequently, $\mathbf{F}_{temporal}^{4}$ is restored to obtain $\mathbf{F}_{st}^{4} \in \mathbb{R}^{B \times T \times C_{4} \times H_{4} \times W_{4}}$ for progressive decoding. Similarly, the third-stage feature $\mathbf{F}^{3}$ is processed by spatial and temporal SSM to produce $\mathbf{F}_{st}^{3}$, which is fused with the upsampled $\mathbf{F}_{st}^{4}$. Repeating this coarse-to-fine process progressively integrates multi-level features, yielding the final spatiotemporal representation $\mathbf{F}_{st} \in \mathbb{R}^{B \times T \times C_{st} \times H \times W}$.


After obtaining the $\mathbf{F}_{st}$, adjacent-frame changes are extracted using absolute differences, yielding $\mathbf{F}_{diff} \in \mathbb{R}^{B \times (T-1) \times C_{d} \times H \times W}$. Subsequently, the $\mathbf{F}_{diff}$ is reshaped by merging the temporal and channel dimensions, enabling it to jointly represent change information across different dynamic footprints. Furthermore, the fusion module comprising convolution and ReLU further enhances the discriminability of the $\mathbf{F}_{diff}$, which is finally used to predict the dynamic footprints. The above process can be formulated as:
\begin{equation}
\label{eq:temporal_difference}
\mathbf{F}_{diff}
=
\mathcal{D}\left(\mathbf{F}_{st}\right)
=
\left[
\left|
\mathbf{F}_{st}^{2}
-
\mathbf{F}_{st}^{1}
\right|,
\ldots,
\left|
\mathbf{F}_{st}^{T}
-
\mathbf{F}_{st}^{T-1}
\right|
\right],
\end{equation}
\begin{equation}
\label{eq:diff_feature_fusion}
\mathbf{F}_{df}
=
\Phi_{df}
\left(
\mathcal{R}_{df}
\left(
\mathbf{F}_{diff}
\right)
\right),
\end{equation}
\begin{equation}
\label{eq:temporal_footprint_prediction}
\hat{\mathbf{Y}}
=
\Phi_{tf}
\left(
\mathbf{F}_{df}
\right)
\end{equation}
where $\mathcal{D}(\cdot)$ denotes the absolute difference operation between adjacent temporal frames, $\mathcal{R}_{\mathrm{df}}(\cdot)$ denotes the feature rearrangement operation that merges the temporal and channel dimensions, $\Phi_{\mathrm{df}}(\cdot)$ denotes the difference feature fusion module, and $\Phi_{\mathrm{tf}}(\cdot)$ denotes the dynamic footprint head.

\section{Experimental Results and Analysis} 
\label{sec:experiment}

\subsection{Data description and preprocessing}
\label{ssec:data}

\subsubsection{Data description}
\label{ssec:data_descrip}
Extensive experiments are conducted on three representative datasets (TSCD~\cite{zhao2024coud}, MUDS~\cite{van2021multi}, and WUSU~\cite{shi2023multi}) to validate the effectiveness of the proposed FootprintNet comprehensively.

\subsubsection{Data preprocessing}
\label{ssec:data_preprocess}
To characterize the dynamic footprints of building changes, labels from adjacent temporal intervals are consolidated into a unified mask, with each class denoting a specific change interval. Taking the TSCD dataset as an example, unchanged regions are labeled as class $0$, while changes occurring between $T_1$ and $T_2$ (i.e., 2016-2018) are assigned class $1$. The remaining single-change intervals are labeled in the same manner. Regions undergoing two or more changes throughout the sequence are assigned an additional multi-change class, denoted as class $N$, indexed by the sequence length $T$ across all datasets. 

Considering the issue of inconsistent sequence lengths in the MUDS dataset, we propose the \textbf{terminal state holding strategy} to pad variable-length sequences. Specifically, we first identify the maximum sequence length among all samples in the dataset. For shorter sequences, the last temporal image is replicated and used as supplementary frames to pad them to a unified length, thereby achieving temporal alignment. This strategy provides a simple yet effective way to align temporal sequences while remaining compatible with the above construction process of temporal change labels, without introducing additional perturbations to the change categories.

\subsection{Experimental setup}
\label{ssec:setup}
\subsubsection{Architecture details}
\label{ssec:architecture}
We adopt the base version of the Mamba-based model~\cite{liu2024vmamba} as the image encoder. The feature maps produced by its four stages are downsampled to $1/4$, $1/8$, $1/16$, and $1/32$ of the original image size, respectively, with the corresponding channel dimensions $C_i$ set to $128$, $256$, $512$, and $1024$. In the ALST branch, both the multi-level feature embedding dimension $C_{sf}$ and temporal fusion dimension $C_{tf}$ are set to $64$. The convolution kernel size is $3\times3$, and the hidden dimension of the action classification head is also set to $64$. In the ATBC branch, the embedding dimension of the feature projection module $C_p$ is set to $128$. Both the spatial Transformer and temporal Transformer contain 1 layer, with the number of attention heads set to $8$ and the dimension of the FFN~\cite{vaswani2017attention} set to $512$. In the SSTS branch, the feature dimension of the spatial SSM and temporal SSM at different stages $C_{st}$ is set to $128$, and the output dimension $C_d$ of the temporal difference fusion module is set to $256$.

\subsubsection{Training details}
\label{ssec:training}

The proposed FootprintNet is implemented based on the PyTorch framework, and model training is conducted on a single NVIDIA RTX 4090 GPU. AdamW~\cite{adamw} is adopted as the optimizer for network parameter optimization, with the initial learning rate set to $2\times 10^{-4}$. Regarding the number of training epochs, the model is trained for $400$ epochs on the TSCD dataset and $200$ epochs on both the MUDS and WUSU datasets. During training, the overall loss function is composed of the losses from ALST, ATBC, and SSTS branches and is defined as follows:
\begin{equation}
\label{eq:overall_loss}
\mathcal{L}_{\mathrm{total}}
=
\lambda_{\mathrm{1}}\mathcal{L}_{\mathrm{SSTS}}
+
\lambda_{\mathrm{2}}\mathcal{L}_{\mathrm{ATBC}}
+
\lambda_{\mathrm{3}}\mathcal{L}_{\mathrm{ALST}}
\end{equation}
where, $\mathcal{L}_{\mathrm{SSTS}}$ denotes the dynamic footprint prediction loss, $\mathcal{L}_{\mathrm{ATBC}}$ denotes the asymmetric temporal-boundary loss, and $\mathcal{L}_{\mathrm{ALST}}$ denotes the action-guided latent state transition loss. $\lambda_{1}$, $\lambda_{2}$, and $\lambda_{3}$ are the corresponding weighting coefficients for these loss terms. Finally, $\lambda_1$, $\lambda_2$, and $\lambda_3$ are set to $1$, $0.1$, and $0.01$, respectively.

\begin{table*}
    \centering
    \caption{Comparison results on the three test sets. The top three results are highlighted in \colorbox{myred}{red}, \colorbox{myorg}{orange}, \colorbox{myyellow}{yellow}. All results are described in percentage (\%).}
    \resizebox{1\textwidth}{!}{
    \begin{tabular}{c|c|c|c}
    \toprule
    \textbf{Method} &
    \multicolumn{1}{c|}{\textbf{TSCD}} &
    \multicolumn{1}{c|}{\textbf{MUDS}} &
    \multicolumn{1}{c}{\textbf{WUSU}} \\
    &
    mPre / mRec / mF1 / mIoU / BCDS &
    mPre / mRec / mF1 / mIoU / BCDS &
    mPre / mRec / mF1 / mIoU / BCDS \\
    \midrule

    BIT$_{22}$~\cite{BIT}
    &
    \colorbox{mywhite}{85.12} /
    \colorbox{mywhite}{78.60} /
    \colorbox{mywhite}{81.68} /
    \colorbox{mywhite}{69.11} /
    \colorbox{mywhite}{80.49}
    &
    \colorbox{mywhite}{25.94} /
    \colorbox{mywhite}{24.57} /
    \colorbox{mywhite}{25.02} /
    \colorbox{mywhite}{14.66} /
    \colorbox{mywhite}{28.96}
    &
    \colorbox{mywhite}{22.35} /
    \colorbox{mywhite}{13.08} /
    \colorbox{mywhite}{15.79} /
    \colorbox{mywhite}{\hphantom{0}9.05} /
    \colorbox{mywhite}{14.54}
    \\
    
    BiFA$_{24}$~\cite{zhang2024bifa}
    &
    \colorbox{mywhite}{88.29} /
    \colorbox{mywhite}{78.54} /
    \colorbox{mywhite}{82.99} /
    \colorbox{mywhite}{71.09} /
    \colorbox{mywhite}{78.01}
    &
    \colorbox{mywhite}{27.29} /
    \colorbox{mywhite}{26.73} /
    \colorbox{mywhite}{26.71} /
    \colorbox{mywhite}{15.84} /
    \colorbox{mywhite}{30.25}
    &
    \colorbox{mywhite}{40.95} /
    \colorbox{mywhite}{22.71} /
    \colorbox{mywhite}{29.07} /
    \colorbox{mywhite}{17.24} /
    \colorbox{mywhite}{25.23}
    \\

    CDMamba$_{25}$~\cite{zhang2025cdmamba}
    &
    \colorbox{mywhite}{84.77} /
    \colorbox{mywhite}{77.42} /
    \colorbox{mywhite}{80.73} /
    \colorbox{mywhite}{67.87} /
    \colorbox{mywhite}{78.58}
    &
    \colorbox{mywhite}{22.99} /
    \colorbox{mywhite}{17.18} /
    \colorbox{mywhite}{18.40} /
    \colorbox{mywhite}{10.26} /
    \colorbox{mywhite}{26.22}
    &
    \colorbox{mywhite}{36.56} /
    \colorbox{mywhite}{25.23} /
    \colorbox{mywhite}{29.68} /
    \colorbox{mywhite}{17.81} /
    \colorbox{myyellow}{28.94}
    \\

    ChangeMamba$_{24}$~\cite{chen2024changemamba}
    &
    \colorbox{mywhite}{90.99} /
    \colorbox{myorg}{85.20} /
    \colorbox{myyellow}{87.86} /
    \colorbox{myyellow}{78.49} /
    \colorbox{myyellow}{84.50}
    &
    \colorbox{myyellow}{29.26} /
    \colorbox{myorg}{33.44} /
    \colorbox{myorg}{30.91} /
    \colorbox{myorg}{18.76} /
    \colorbox{myred}{\textbf{37.78}}
    &
    \colorbox{myred}{\textbf{54.54}} /
    \colorbox{mywhite}{22.47} /
    \colorbox{myorg}{31.68} /
    \colorbox{myorg}{19.08} /
    \colorbox{mywhite}{25.34}
    \\

    FoBa$_{25}$~\cite{zhang2025foba}
    &
    \colorbox{myyellow}{92.52} /
    \colorbox{myyellow}{85.14} /
    \colorbox{myorg}{88.59} /
    \colorbox{myorg}{79.67} /
    \colorbox{myorg}{84.49}
    &
    \colorbox{myorg}{31.15} /
    \colorbox{myyellow}{28.43} /
    \colorbox{myyellow}{29.63} /
    \colorbox{myyellow}{17.85} /
    \colorbox{mywhite}{29.13}
    &
    \colorbox{myorg}{51.30} /
    \colorbox{mywhite}{22.44} /
    \colorbox{myyellow}{31.05} /
    \colorbox{mywhite}{18.61} /
    \colorbox{mywhite}{24.97}
    \\

    CUCD$_{26}$~\cite{hafner2025continuous}
    &
    \colorbox{mywhite}{90.68} /
    \colorbox{mywhite}{80.77} /
    \colorbox{mywhite}{85.28} /
    \colorbox{mywhite}{74.68} /
    \colorbox{mywhite}{78.85}
    &
    \colorbox{mywhite}{22.57} /
    \colorbox{mywhite}{24.75} /
    \colorbox{mywhite}{22.89} /
    \colorbox{mywhite}{13.32} /
    \colorbox{mywhite}{28.06}
    &
    \colorbox{mywhite}{39.14} /
    \colorbox{myyellow}{25.59} /
    \colorbox{mywhite}{30.88} /
    \colorbox{myyellow}{18.77} /
    \colorbox{mywhite}{27.22}
    \\

    CAIMNet$_{26}$~\cite{CAIMNet}
    &
    \colorbox{mywhite}{73.97} /
    \colorbox{mywhite}{57.76} /
    \colorbox{mywhite}{64.82} /
    \colorbox{mywhite}{47.96} /
    \colorbox{mywhite}{63.02}
    &
    \colorbox{mywhite}{23.49} /
    \colorbox{mywhite}{15.08} /
    \colorbox{mywhite}{17.96} /
    \colorbox{mywhite}{10.14} /
    \colorbox{mywhite}{18.16}
    &
    \colorbox{mywhite}{33.42} /
    \colorbox{myorg}{27.94} /
    \colorbox{mywhite}{29.19} /
    \colorbox{mywhite}{17.47} /
    \colorbox{myorg}{31.51}
    \\

    RSSM$_{26}$~\cite{zhu2026rsssm}
    &
    \colorbox{myred}{\textbf{93.81}} /
    \colorbox{mywhite}{69.57} /
    \colorbox{mywhite}{79.47} /
    \colorbox{mywhite}{66.39} /
    \colorbox{mywhite}{66.69}
    &
    \colorbox{mywhite}{27.04} /
    \colorbox{mywhite}{27.28} /
    \colorbox{mywhite}{26.79} /
    \colorbox{mywhite}{15.87} /
    \colorbox{myyellow}{32.03}
    &
    \colorbox{mywhite}{17.66} /
    \colorbox{mywhite}{\hphantom{0}3.92} /
    \colorbox{mywhite}{\hphantom{0}6.32} /
    \colorbox{mywhite}{\hphantom{0}3.32} /
    \colorbox{mywhite}{\hphantom{0}4.38}
    \\

    \midrule

    FootprintNet
    &
    \colorbox{myorg}{92.28} /
    \colorbox{myred}{\textbf{86.09}} /
    \colorbox{myred}{\textbf{89.01}} /
    \colorbox{myred}{\textbf{80.34}} /
    \colorbox{myred}{\textbf{85.57}}
    &
    \colorbox{myred}{\textbf{34.99}} /
    \colorbox{myred}{\textbf{31.44}} /
    \colorbox{myred}{\textbf{32.56}} /
    \colorbox{myred}{\textbf{19.88}} /
    \colorbox{myorg}{35.18}
    &
    \colorbox{myyellow}{49.76} /
    \colorbox{myred}{\textbf{32.48}} /
    \colorbox{myred}{\textbf{39.24}} /
    \colorbox{myred}{\textbf{24.92}} /
    \colorbox{myred}{\textbf{35.42}}
    \\

    \bottomrule
    \end{tabular}
    }
    \label{tab:comparison_sotas}
\end{table*}

\subsubsection{Evaluation metrics}
\label{ssec:evaluation}
We adopt widely used evaluation metrics, including precision (mPre), recall (mRec), F1 score (mF1), and mean Intersection over Union (mIoU) to quantitatively assess model performance.

\textbf{Building Change Dynamics Score:} In addition, the aforementioned mF1 and mIoU generally treat all categories as mutually independent prediction targets and assign equal penalties to misclassifications among different categories, making it difficult to reflect the temporal proximity between the prediction and the ground-truth dynamic footprint. To address this limitation, we propose the building change dynamics score (BCDS), which evaluates predictions from the perspective of semantic preservation in building change dynamics. 

The core idea is that even when the prediction does not exactly match the label, differentiated scores can still be assigned according to the temporal relationship between the predicted and ground-truth categories. Specifically, for single-change categories, we evaluate the prediction quality from three aspects: change existence, i.e., whether the model can correctly identify the occurrence of building change. Single-change attribute, i.e., whether the model can distinguish a single change event from a repeated change process. And single-change temporal distance, i.e., whether the model can accurately locate the dynamics footprint and assign scores according to the degree of temporal offset. Finally, the three scores are equally averaged to obtain the comprehensive score for single-change categories. The calculation process can be formulated as:
\begin{equation}
\label{eq:bcds_category_set}
\mathcal{C}_{\mathrm{s}}=\{1,2,\ldots,K-1\},
\quad
\mathcal{C}_{\mathrm{m}}=\{K\},
\end{equation}

\begin{equation}
\label{eq:bcds_single_components}
\begin{aligned}
S_i^{\mathrm{chg}}
&=
\mathbb{I}\left(\hat{y}_i \in \mathcal{C}_{\mathrm{s}} \cup \mathcal{C}_{\mathrm{m}}\right), \\
S_i^{\mathrm{attr}}
&=
\mathbb{I}\left(\hat{y}_i \in \mathcal{C}_{\mathrm{s}}\right), \\
S_i^{\mathrm{temp}}
&=
\mathbb{I}\left(\hat{y}_i \in \mathcal{C}_{\mathrm{s}}\right)
\left(
1-
\frac{
\left|\hat{y}_i-y_i\right|
}{
K-2
}
\right),
\quad y_i\in\mathcal{C}_{\mathrm{s}}
\end{aligned}
\end{equation}

\begin{equation}
\label{eq:bcds_single_score}
H_i^{\mathrm{single}}
=
\frac{
S_i^{\mathrm{chg}}
+
S_i^{\mathrm{attr}}
+
S_i^{\mathrm{temp}}
}{3},
\quad
y_i\in\mathcal{C}_{\mathrm{s}}
\end{equation}
where $\mathbb{I}(\cdot)$ denotes the indicator function, which equals $1$ when the specified condition is satisfied and $0$ otherwise. $\mathcal{C}_{\mathrm{s}}$ denotes the set of single-change categories, while $\mathcal{C}_{\mathrm{m}}$ denotes the set of multi-change categories. $S_i^{\mathrm{chg}}$, $S_i^{\mathrm{attr}}$, and $S_i^{\mathrm{temp}}$ represent the change-existence score, single-change attribute score, and temporal-distance score, respectively.

For the multi-change category, we follow the same principle and decompose its score into change existence and whether the prediction corresponds to a multi-change process. The score for the multi-change category is defined as:

\begin{equation}
\label{eq:bcds_multi_components}
\begin{aligned}
S_i^{\mathrm{chg}}
&=
\mathbb{I}
\left(
\hat{y}_i \in \mathcal{C}_{\mathrm{s}} \cup \mathcal{C}_{\mathrm{m}}
\right), \\
S_i^{\mathrm{mul}}
&=
\mathbb{I}
\left(
\hat{y}_i \in \mathcal{C}_{\mathrm{m}}
\right),
\quad
y_i \in \mathcal{C}_{\mathrm{m}}
\end{aligned}
\end{equation}

\begin{equation}
\label{eq:bcds_multi_score}
H_i^{\mathrm{multi}}
=
\frac{
S_i^{\mathrm{chg}}
+
S_i^{\mathrm{mul}}
}{2},
\quad
y_i \in \mathcal{C}_{\mathrm{m}}
\end{equation}

It is worth noting that, to avoid metric bias caused by class imbalance, we adopt macro-averaging to ensure that different change categories contribute equally during evaluation. Meanwhile, unchanged regions are excluded from the metric calculation, allowing the evaluation to focus more specifically on the building change process. Finally, the single-change category score and the multi-change category score are weighted and averaged to obtain the BCDS.
\begin{equation}
\label{eq:bcds}
\mathrm{BCDS}
=
\frac{
H_{\mathrm{single}}
+
H_{\mathrm{multi}}
}{2}
\end{equation}




\begin{table}[t]
    \centering
    \caption{Class-wise IoU comparison results on the TSCD test set.}
    \label{tab:comparison_class_wise_tscd}

    \resizebox{\columnwidth}{!}{
    \begin{tabular}{l c c c c}
        \toprule
        \multirow{2}{*}{\textbf{Method}} &
        \multicolumn{4}{c}{\textbf{Class-wise IoU}} \\
        \cmidrule(lr){2-5}
        &
        \makecell{\textbf{C$_1$}\\$T_1$--$T_2$} &
        \makecell{\textbf{C$_2$}\\$T_2$--$T_3$} &
        \makecell{\textbf{C$_3$}\\$T_3$--$T_4$} &
        \makecell{\textbf{C$_4$}\\Multi-change} \\
        \midrule

        BIT$_{22}$
        & 70.23 & 68.34 & 74.03 & 63.85 \\

        BiFA$_{24}$
        & 70.85 & 75.16 & 75.86 & 62.49 \\

        CDMamba$_{25}$ 
        & 65.10 & 71.81 & 74.42 & 60.17 \\

        ChangeMamba$_{24}$ 
        & 78.26 & 83.07 & 81.99 & 70.65 \\

        FoBa$_{25}$ 
        & 80.37 & \textbf{84.30} & 83.07 & 70.95 \\
        
        CUCD$_{26}$ 
        & 74.19 & 83.35 & 78.40 & 62.77 \\
        
        CAIMNet$_{26}$
        & 47.09 & 48.51 & 51.09 & 46.07 \\
        
        RSSM$_{26}$ 
        & 69.27 & 77.22 & 65.60 & 53.46 \\

        \midrule
        FootprintNet
        & \textbf{82.43}
        & 83.37
        & \textbf{84.13}
        & \textbf{71.44} \\

        \bottomrule
    \end{tabular}
    }
\end{table}

\begin{table*}
    \centering
    \caption{Ablation study on Different Branches.}
    \resizebox{0.85\textwidth}{!}{
    \begin{tabular}{c|c|c|c|c}
    \toprule
    \multicolumn{1}{c|}{} &
    \multicolumn{1}{c|}{} &
    \multicolumn{1}{c|}{} &
    \multicolumn{1}{c|}{\textbf{TSCD}} &
    \multicolumn{1}{c}{\textbf{MUDS}} \\
    ALST & ATBC & SSTS &
    mPre / mRec / mF1 / mIoU / BCDS &
    mPre / mRec / mF1 / mIoU / BCDS \\
    \midrule

    {$\times$}
    & {$\times$}
    & {$\times$}
    & 90.99 / 85.20 / 87.86 / 78.49 / 84.50
    & 29.26 / 33.44 / 30.91 / 18.76 / 37.78 \\

    \midrule

    $\checkmark$
    & {$\times$}
    & {$\times$}
    & 93.79 / 83.30 / 88.08 / 78.95 / 81.88
    & 29.68 / 34.08 / 31.62 / 19.29 / 37.08 \\

    {$\times$}
    & $\checkmark$
    & {$\times$}
    & 93.32 / 83.97 / 88.17 / 79.13 / 81.79
    & 32.69 / 30.92 / 31.55 / 19.18 / 33.42 \\

    {$\times$}
    & {$\times$}
    & $\checkmark$
    & 90.60 / 85.82 / 88.02 / 78.71 / 85.28
    & 33.13 / 29.50 / 31.05 / 18.89 / 28.74 \\

    $\checkmark$
    & $\checkmark$
    & {$\times$}
    & 94.09 / 84.41 / 88.90 / 80.22 / 83.58
    & 31.63 / 33.90 / 31.98 / 19.40 / 38.04 \\

    $\checkmark$
    & {$\times$}
    & $\checkmark$
    & 90.99 / 85.82 / 88.21 / 79.00 / 85.60
    & 32.88 / 31.82 / 31.94 / 19.39 / 35.44 \\

    {$\times$}
    & $\checkmark$
    & $\checkmark$
    & 93.57 / 83.98 / 88.44 / 79.49 / 83.01
    & 31.57 / 32.58 / 31.84 / 19.36 / 37.30 \\

    \midrule

    $\checkmark$
    & $\checkmark$
    & $\checkmark$
    & 92.28 / 86.09 / 89.01 / 80.34 / 85.57
    & 34.99 / 31.44 / 32.56 / 19.88 / 35.18 \\

    \bottomrule
    \end{tabular}
    }
    \label{tab:abalation_compnents}
\end{table*}

\subsection{Performance comparison}
\label{ssec:performance}


To evaluate the effectiveness of FootprintNet, we compare it with representative state-of-the-art methods, including CNN-based CAIMNet~\cite{CAIMNet}, Transformer-based BIT~\cite{BIT}, BiFA~\cite{zhang2024bifa}, and CUCD~\cite{hafner2025continuous}, as well as Mamba-based ChangeMamba~\cite{chen2024changemamba}, CDMamba~\cite{zhang2025cdmamba}, FoBa~\cite{zhang2025foba}, and RSSM~\cite{zhu2026rsssm}. Since some methods are not directly applicable to UBDD, we introduce a unified temporal difference processing module to adapt them to temporal inputs and dynamic footprint prediction. For a fair comparison, all competing methods are implemented using their official PyTorch codes and retrained under identical experimental settings.

\subsubsection{Quantitative results}
\label{ssec:quantitative}




Table~\ref{tab:comparison_sotas} reports the quantitative results on the TSCD, MUDS, and WUSU test sets, where \colorbox{myred}{red}, \colorbox{myorg}{orange}, and \colorbox{myyellow}{yellow} denote the best, second-best, and third-best results, respectively. Obviously, FootprintNet consistently outperforms representative Transformer-based methods, including BIT, BiFA, and CUCD. Compared with BIT and BiFA, FootprintNet improves $mF1$ by 7.33\%/6.02\%, 7.54\%/5.85\%, and 23.45\%/10.17\% on TSCD, MUDS, and WUSU, respectively. Notably, compared with CUCD, which is originally designed for temporal change detection, FootprintNet achieves gains of 5.66\%/6.72\%, 6.76\%/7.12\%, and 5.52\%/8.20\% in terms of $mIoU$/BCDS on the three datasets, respectively. These results demonstrate the effectiveness of FootprintNet for UBDD and its superiority in characterizing the dynamic footprints of building changes.

Compared with Mamba-based methods, FootprintNet also shows clear advantages. It outperforms CDMamba and FoBa across all metrics, achieving $mIoU$ gains of 12.47\%/0.67\%, 9.62\%/2.03\%, and 7.11\%/6.31\% on TSCD, MUDS, and WUSU, respectively. Compared with RS-SSM, FootprintNet yields slightly lower $mPre$ but higher $mRec$, $mF1$, $mIoU$, and BCDS, indicating a more balanced detection capability. Although FootprintNet obtains higher $mIoU$ than ChangeMamba, its lower BCDS on MUDS suggests that temporal-offset errors remain in dynamic footprint characterization.

In addition, we quantitatively evaluate the class-wise performance to assess the effectiveness of FootprintNet across different dynamic footprint categories, as reported in Table~\ref{tab:comparison_class_wise_tscd}. FootprintNet exhibits more balanced advantages across the four change categories. Compared with the second-best method, FoBa, it improves the scores for the single-change categories $C_1$ and $C_3$ by 2.06\% and 1.06\%, respectively. Although its performance on $C_2$ is slightly lower than that of FoBa, it still outperforms other competing methods, such as ChangeMamba. For the multi-change category $C_4$, which better reflects the model's understanding of temporal change processes, FootprintNet achieves a 0.49\% improvement over FoBa, further demonstrating its ability to characterize complex dynamic footprints of building changes.

Overall, the quantitative results verify the contributions of ALST, ATBC, and SSTS to UBDD. By integrating state-transition priors with explicit change-boundary guidance, FootprintNet learns more discriminative spatiotemporal representations, thereby improving the characterization of dynamic building-change footprints.

\begin{figure*}
    \centering
    \includegraphics[width=0.99\textwidth]{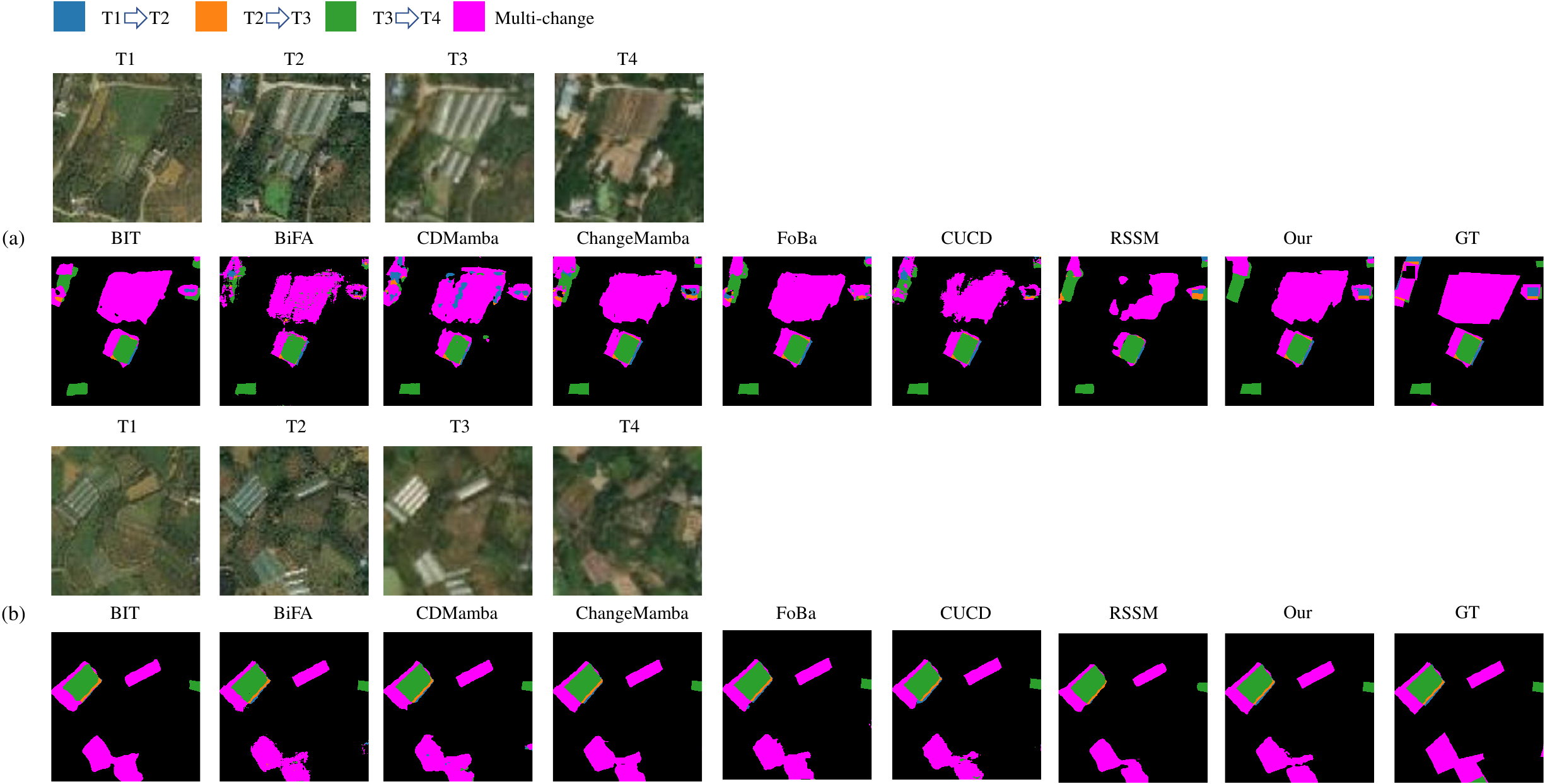}
    \caption{Visualization results of different methods on the TSCD test set. The black denotes unchanged areas.}
    \label{fig:vis_tscd}
\end{figure*}

\subsubsection{Qualitative results}
\label{ssec:qualitative}

\begin{figure*}[!t]
    \centering
    \includegraphics[width=0.99\textwidth]{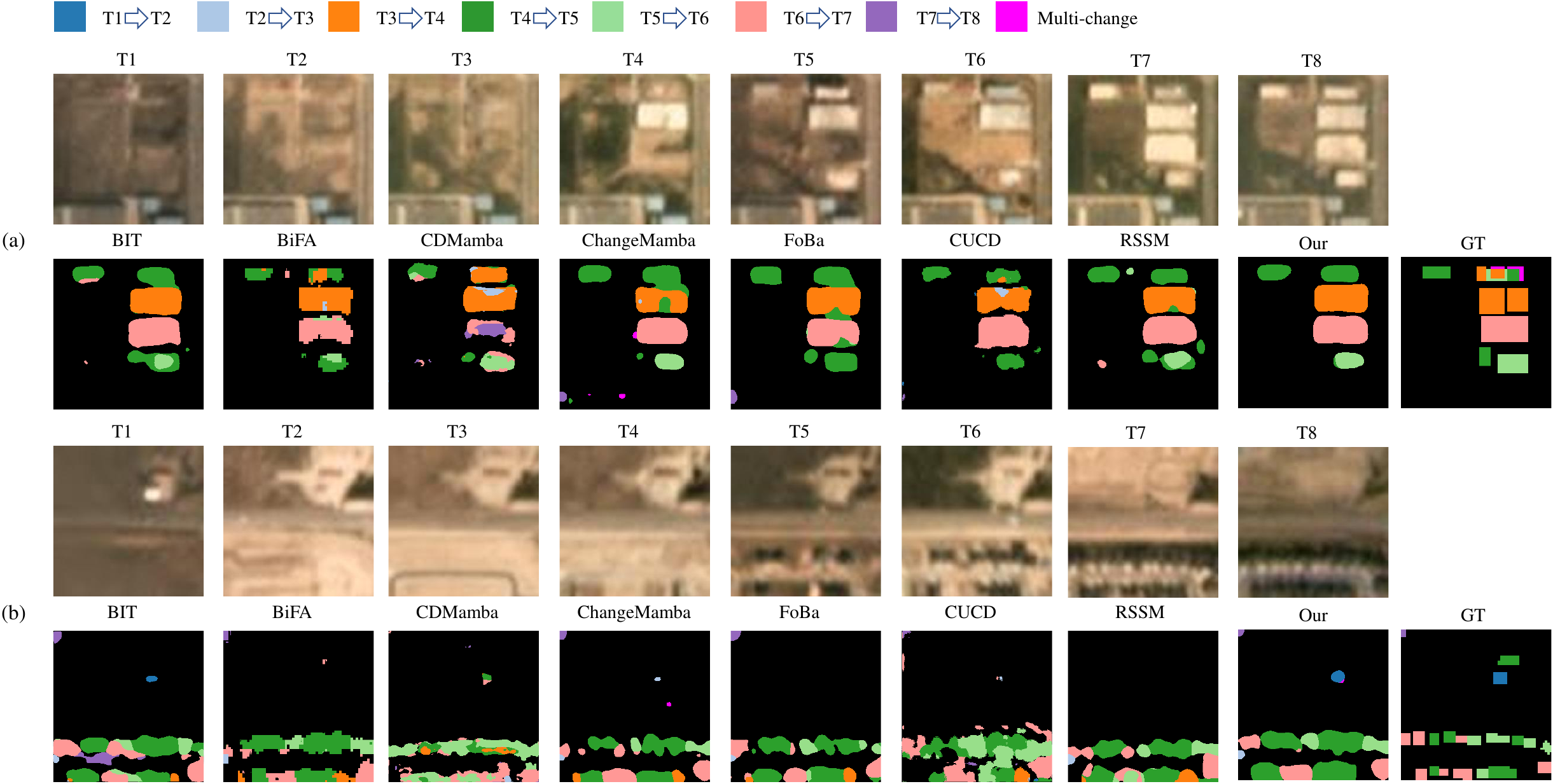}
    \vspace{0.7mm}
    \caption{Visualization results of different methods on the MUDS test set. 
    The black regions denote unchanged areas.}
    \label{fig:vis_muds}
    \vspace{1.5mm}
\end{figure*}

\begin{figure*}[!t]
    \centering
    \includegraphics[width=0.99\textwidth]{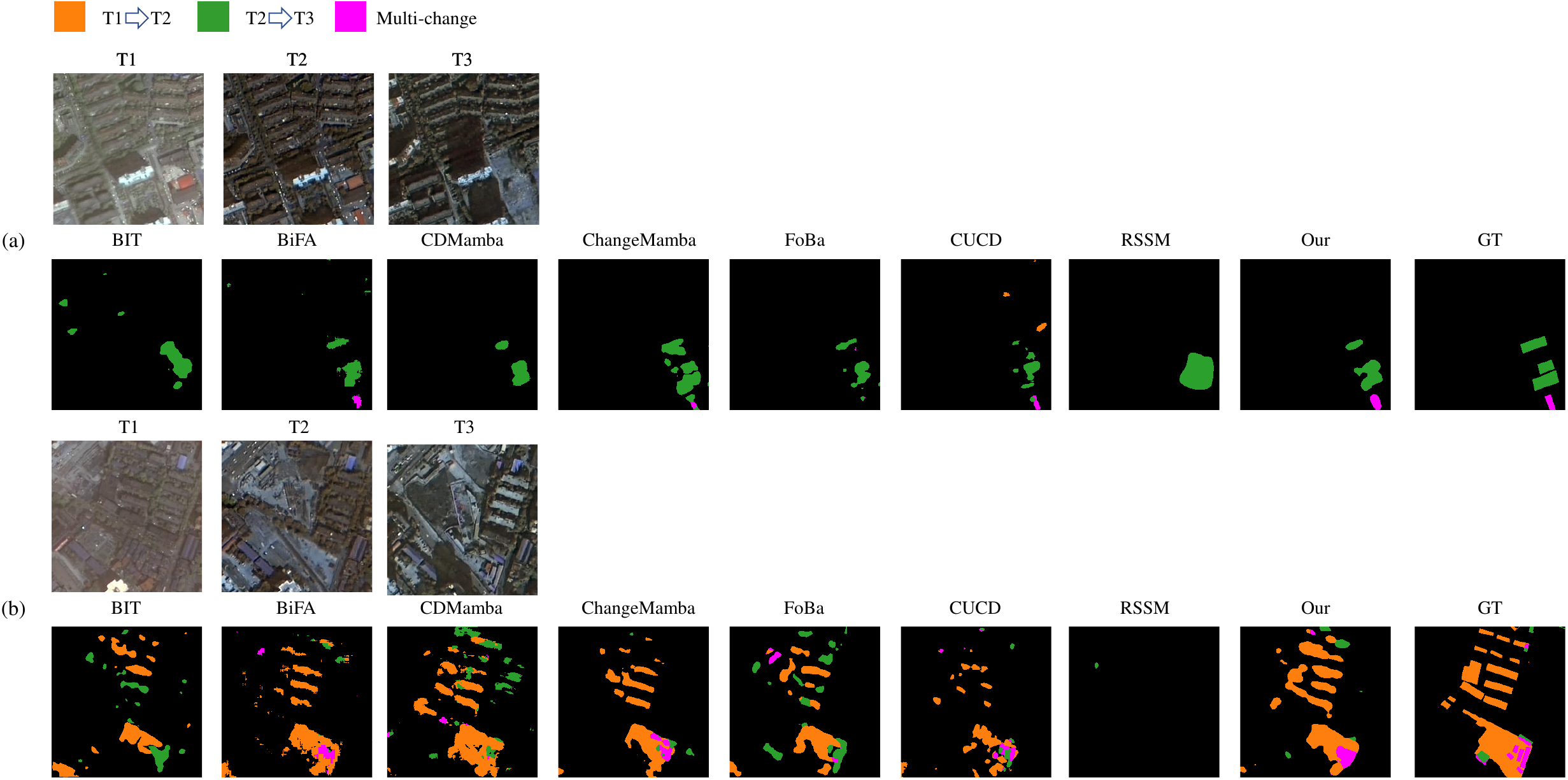}
    \vspace{0.7mm}
    \caption{Visualization results of different methods on the WUSU test set. 
    The black regions denote unchanged areas.}
    \label{fig:vis_wusu}
    \vspace{1mm}
\end{figure*}

To further validate the effectiveness of the proposed method, we conduct qualitative analyses on TSCD, MUDS, and WUSU, as illustrated in Fig. ~\ref{fig:vis_tscd}-~\ref{fig:vis_wusu}.

Visualization on TSCD (Fig.~\ref{fig:vis_tscd}): Fig.~\ref{fig:vis_tscd}(a)-(b) presents representative qualitative comparisons. In Fig.~\ref{fig:vis_tscd}(a), although most competing methods roughly localize the large-scale multi-change regions, FootprintNet produces changed regions and dynamic footprints that are more consistent with the label, especially by suppressing false detections in the upper-right region. In Fig.~\ref{fig:vis_tscd}(b), FootprintNet better preserves the geometric structure and boundary details of multi-change regions. 


Visualization on MUDS (Fig. ~\ref{fig:vis_muds}): Similarly, we selected several representative samples from the MUDS dataset for illustration, as shown in Fig. ~\ref{fig:vis_muds} (a)-(b). In the  Fig.~\ref{fig:vis_muds}(a), FootprintNet more accurately localizes changed regions and identifies dynamic footprints, while FoBa, ChangeMamba, and RS-SSM produce evident false detections. Compared with BIT, FootprintNet also preserves clearer building boundaries with fewer adhesions between adjacent objects. In the small building-change scenario of Fig.~\ref{fig:vis_muds}(b), FootprintNet still provides more accurate localization despite missed detections by most methods. 

Visualization on WUSU (Fig. \ref{fig:vis_wusu}): Fig.~\ref{fig:vis_wusu}(a)-(b) presents representative visual comparisons on the WUSU dataset. In the Fig.~\ref{fig:vis_wusu}(a), FootprintNet suppresses scattered false detections, avoids the missed objects of CDMamba and the excessive responses of RS-SSM, and achieves more complete localization and accurate dynamic footprint recognition. In the complex multi-object scene of Fig.~\ref{fig:vis_wusu}(b), it reduces false detections, and temporal-class confusion, thereby more faithfully recovering the spatial distribution and dynamic footprints of changed buildings.



\begin{figure*}
    \centering
    \includegraphics[width=0.99\textwidth]{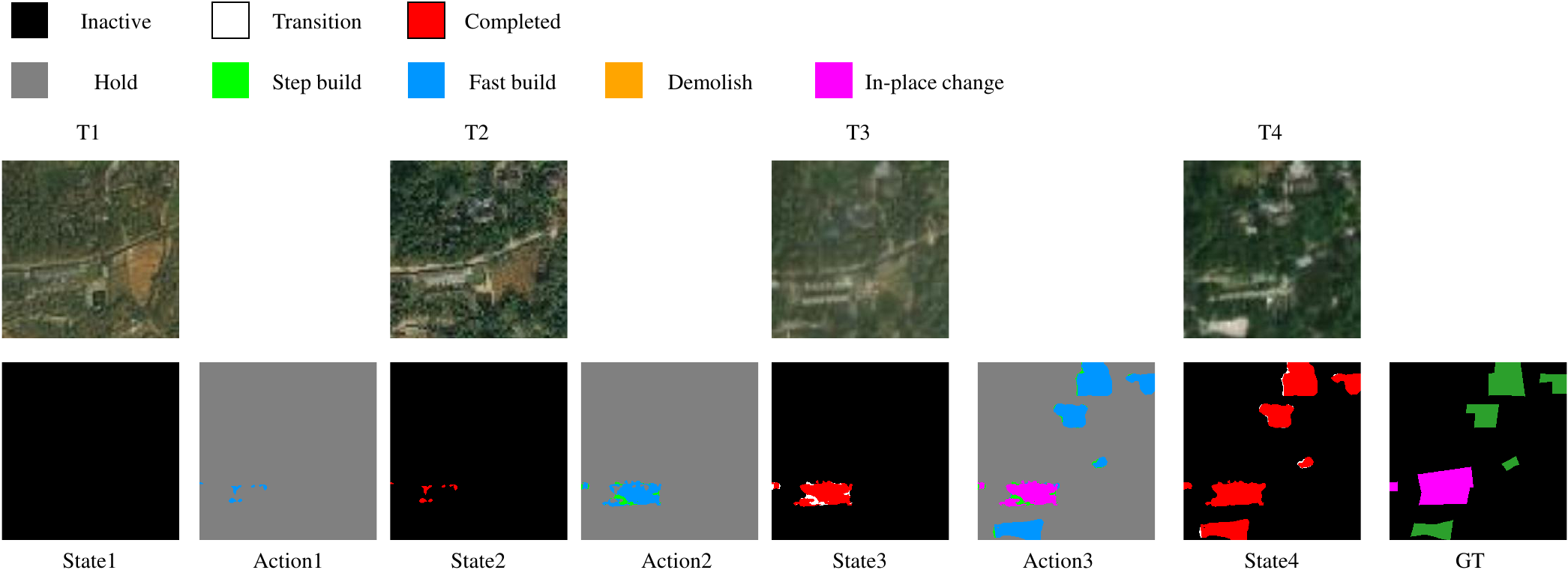}
    \caption{Visualization of state-action trajectories across different time steps on the TSCD dataset. T1-T4 denote the multi-temporal images, while State1-State4 represent their corresponding state maps. Action1-Action3 indicate the actions executed to drive state transitions between adjacent temporal.
    }
    \label{fig:state_action}
\end{figure*}

\begin{figure*}
    \centering
    \includegraphics[width=0.99\textwidth]{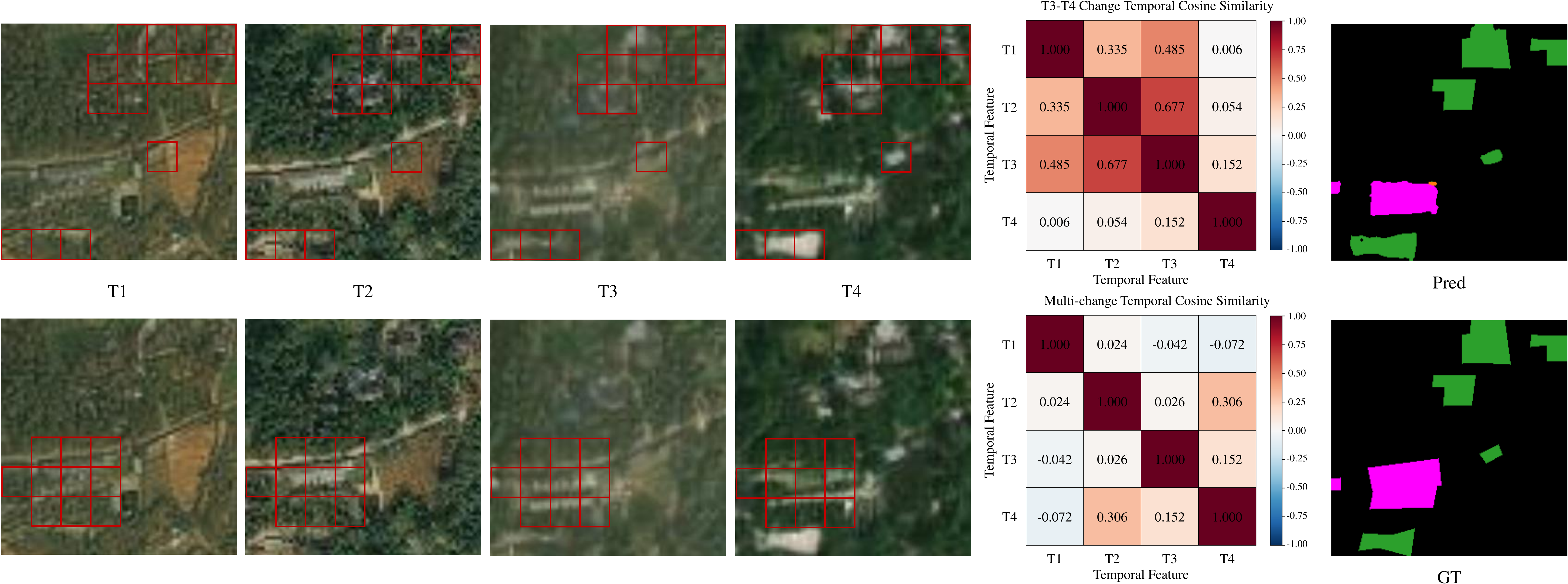}
    \caption{Visualization of feature similarities across different dynamic footprints. The red boxes in T1-T4 indicate the selected image patches. For each temporal, the features within each selected patch are averaged, and the resulting representations are used to compute the cosine similarity maps.
    }
    \label{fig:vis_cars}
\end{figure*}

\subsection{Ablation studies}
\label{ssec:ablation}


In this subsection, we conduct a series of experiments on the TSCD and WUSU datasets to evaluate the impact of each branch on model performance, with detailed results presented in Table~\ref{tab:abalation_compnents}--\ref{tab:ablation_layers}.

\subsubsection{Effects of Different Branches in FootprintNet}
\label{ssec: Different Components}



To evaluate the contribution of each branch to detection performance, we conduct eight ablation experiments on the TSCD and MUDS datasets. Specifically, ChangeMamba is adopted as the baseline, upon which the ALST, ATBC, and SSTS branches are progressively incorporated. As reported in Table~\ref{tab:abalation_compnents}, introducing these branches either individually or in combination consistently improves the overall performance over the baseline on both datasets. It is observed that FootprintNet improves $mF1$ by 1.15\% and 1.85\%, and $mIoU$ by 1.65\% and 1.12\% on TSCD and MUDS, respectively. Moreover, BCDS increases by 1.07\% on TSCD but decreases slightly on MUDS, indicating that FootprintNet remains marginally inferior to the baseline in preserving change semantics on this dataset. Overall, these results validate the effectiveness of the ALST, ATBC, and SSTS branches.

Notably, the dynamic footprint detection performance of FootprintNet is significantly improved after introducing the ALST and ATBC branches. These quantitative results demonstrate that integrating state-transition priors into model constraints and leveraging temporal-boundary guidance to distinguish different temporal-dynamics regions in the feature space are crucial for enhancing the model’s understanding of temporal change processes. To further provide an intuitive analysis of their effectiveness, we visualize the state-action predictions at different time steps on the TSCD, as shown in Fig.~\ref{fig:state_action}. In the state maps, black, white, and red denote the Inactive, Transition, and Completed states, respectively. In the action maps, gray, green, blue, orange, and purple denote the Hold, Step build, Fast build, Demolish, and In-place change actions, respectively. As observed, the building region in the central part of the image undergoes fast build from $T_2$ to $T_3$, followed by an in-place change in the subsequent temporal observation. Meanwhile, multiple building regions exhibit construction processes from $T_3$ to $T_4$. The resulting state-action trajectories are highly consistent with the actual change processes reflected in the multi-temporal images from $T_1$ to $T_4$, intuitively demonstrating that ALST can effectively characterize the dynamics footprints.

In addition, Fig.~\ref{fig:vis_cars} visualizes the cross-temporal feature similarities of image patches corresponding to changed regions. In the similarity map for the $T_3$--$T_4$ change category, the patch features from the first three temporal observations exhibit high similarity, whereas a clear similarity boundary emerges at $T_4$. A similar pattern can also be observed in the patch-similarity map of the multi-change category. These results intuitively demonstrate that ATBC can effectively separate different temporal change patterns in the feature space and enhance the model’s discrimination of change boundaries.

        

\begin{table}
    \centering
    \caption{Effect of the different dims}
    \resizebox{0.4\textwidth}{!}{
    \begin{tabular}{c|c}
    \toprule
    \multicolumn{1}{c|}{} &
    \multicolumn{1}{c}{\textbf{MUDS}} \\
    Dims & mPre / mRec / mF1 / mIoU / BCDS \\
        
    \midrule
    16
    & 32.86 / 32.14 / 32.22 / 19.58 /  36.52 \\
    32
    & 36.93 / 29.70 / 32.49 / 19.81 / 32.60 \\
    64
    & 34.99 / 31.44 / 32.56 / 19.88 / 35.18 \\
    128
    & 31.32 / 33.24 / 31.91 / 19.54 / 35.26\\
    256
    & 30.61 / 32.91 / 31.60 / 19.23 / 34.78 \\
    \bottomrule
    \end{tabular}
    }
    \label{tab:ablation_dims}
\end{table}

\begin{table}
    \centering
    \caption{Effect of the different layers}
    \resizebox{0.4\textwidth}{!}{
    \begin{tabular}{c|c}
    \toprule
    \multicolumn{1}{c|}{} &
    \multicolumn{1}{c}{\textbf{MUDS}} \\
    Layers & mPre / mRec / mF1 / mIoU / BCDS \\
        
    \midrule
    1
    & 34.99 / 31.44 / 32.56 / 19.88 / 35.18 \\
    2
    & 33.43 / 32.15 / 32.45 / 19.86 / 32.35 \\
    4
    & 31.83 / 33.25 / 31.96 / 19.50 / 34.27 \\
    8
    & 33.25 / 31.49 / 31.78 / 19.28 / 36.38\\
    16
    & 38.80 / 26.81 / 31.45 / 19.11 / 27.33\\
    \bottomrule
    \end{tabular}
    }
    \label{tab:ablation_layers}
\end{table}

\subsubsection{Different dims in ALST} 
To investigate the effect of the feature dimension in the ALST branch on model performance, we conduct experiments on the MUDS dataset, as reported in Table~\ref{tab:ablation_dims}. The performance generally improves as the feature dimension increases from $16$, reaching the highest $mIoU$ at a dimension of $64$. Further increasing the feature dimension, however, leads to performance degradation. Therefore, the feature dimension of the ALST branch is set to $64$.

\subsubsection{Different layers in ATBC}

To investigate the effect of the number of Transformer layers in the ATBC branch, we conduct experiments on the MUDS dataset, as reported in Table~\ref{tab:ablation_layers}. The overall performance gradually decreases as the numbers of spatial and temporal Transformer layers are simultaneously increased. This may be because ATBC operates on high-level semantic features, for which shallow Transformer structures are already sufficient to model spatial and temporal dependencies. Increasing the network depth may lead to feature over-smoothing, thereby weakening temporal-boundary responses across different change stages. Therefore, both the spatial and temporal Transformers in ATBC are set to 1.

\section{Conclusion}
\label{sec:conclusion}


In this paper, we extend conventional MTCD and introduce Urban Building Dynamics Detection (UBDD), which generalizes single-change-time localization to the unified identification of single- and multi-change processes, thereby relaxing the implicit single-change assumption of existing TCD methods. We further propose FootprintNet, which abstracts building-change processes as interactions between latent states and actions and imposes state-action transition constraints to guide the learning of causally coherent change trajectories. Meanwhile, temporal change-boundary cues are also incorporated to enhance feature discrimination across different dynamic stages, enabling accurate characterization of dynamic building-change footprints. In addition, we introduce the Building Change Dynamics Score (BCDS), which assigns differentiated scores according to the degree of semantic preservation and the temporal offset between predictions and labels. It addresses the limitation of conventional metrics that treat individual classes as independent targets and therefore fail to reflect the temporal proximity between predicted footprints and their corresponding labels. Extensive experiments on three public datasets demonstrate that the proposed method yields more accurate predictions of building-change dynamic footprints.





\ifCLASSOPTIONcaptionsoff
  \newpage
\fi

{\small
\bibliographystyle{IEEEtran}
\bibliography{refs}
}


\end{document}